\documentclass[sigconf]{acmart}
\AtBeginDocument{%
  \providecommand\BibTeX{{%
    \normalfont B\kern-0.5em{\scshape i\kern-0.25em b}\kern-0.8em\TeX}}}

\usepackage{graphicx}
\usepackage{subfigure}
\usepackage{multirow}

\copyrightyear{2022}
\acmYear{2022}
\setcopyright{acmcopyright}
\acmConference[MM '22]{Proceedings of the 30th ACM
International Conference on Multimedia}{October 10--14, 2022}{Lisboa, Portugal}
\acmBooktitle{Proceedings of the 30th ACM International Conference on Multimedia
(MM '22), October 10--14, 2022, Lisboa, Portugal}
\acmPrice{15.00}
\acmDOI{10.1145/3503161.3548029}
\acmISBN{978-1-4503-9203-7/22/10}

\begin{document}

\title{Estimation of Reliable Proposal Quality for Temporal Action Detection}

\author{Junshan Hu}
\authornote{Work done during an internship at Alibaba Inc.}
\email{junshan@mail.ustc.edu.cn}
\affiliation{%
  \institution{University of Science and Technology of China}
  \country{}
}

\author{Chaoxu Guo}
\email{chaoxu.gcx@alibaba-inc.com}
\affiliation{%
  \institution{Alibaba Group}
  \country{}
}

\author{Liansheng Zhuang}
\authornote{Corresponding Author}
\email{lszhuang@ustc.edu.cn}
\affiliation{%
  \institution{University of Science and Technology of China}
  \country{}
}

\author{Biao Wang}
\email{eric.wb@alibaba-inc.com}
\affiliation{%
  \institution{Alibaba Group}
  \country{}
}

\author{Tiezheng Ge}
\email{tiezheng.gtz@alibaba-inc.com}
\affiliation{%
  \institution{Alibaba Group}
  \country{}
}

\author{Yuning Jiang}
\email{mengzhu.jyn@alibaba-inc.com}
\affiliation{%
  \institution{Alibaba Group}
  \country{}
}

\author{Houqiang Li}
\email{lihq@ustc.edu.cn}
\affiliation{%
  \institution{University of Science and Technology of China}
  \country{}
}

\renewcommand{\shortauthors}{Junshan Hu et al.}

\begin{abstract}
Temporal action detection (TAD) aims to locate and recognize the actions in an untrimmed video. 
Anchor-free methods have made remarkable progress which mainly formulate TAD into two tasks: classification and localization using two separate branches. 
This paper reveals the temporal misalignment between the two tasks hindering further progress.
To address this, we propose a new method that gives insights into moment and region perspectives simultaneously to align the two tasks by acquiring reliable proposal quality.
For the moment perspective, Boundary Evaluate Module (BEM) is designed which focuses on local appearance and motion evolvement to estimate boundary quality and adopts a multi-scale manner to deal with varied action durations.
For the region perspective, we introduce Region Evaluate Module (REM) which uses a new and efficient sampling method for proposal feature representation containing more contextual information compared with point feature to refine category score and proposal boundary.
The proposed \textbf{B}oundary Evaluate Module and \textbf{R}egion \textbf{E}valuate \textbf{M}odule (BREM) are generic, and they can be easily integrated with other anchor-free TAD methods to achieve superior performance.
In our experiments, BREM is combined with two different frameworks and improves the performance on THUMOS14 by 3.6$\%$ and 1.0$\%$ respectively, reaching a new state-of-the-art (63.6$\%$ average $m$AP). 
Meanwhile, a competitive result of 36.2\% average $m$AP is achieved on ActivityNet-1.3 with the consistent improvement of BREM. The codes are released at \url{https://github.com/Junshan233/BREM}.
\end{abstract}




\maketitle

\begin{figure}[h]
  \vspace{-0.3cm}
  \centering
  \includegraphics[width=\linewidth]{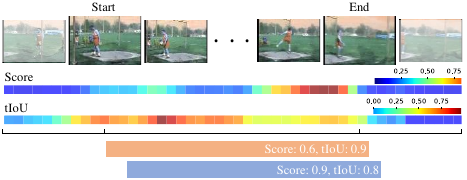}
  \caption{Illustration of misaligned temporal distribution between classification score (\textit{Score}) and localization quality (\textit{tIoU}). Orange and blue blocks indicate two proposals.}
  \label{fig:motivation_a}
  \Description{The figure shows the classification score and localization quality of each frame of a video, }  
  \vspace{-0.3cm}
\end{figure}
                                                                                  
\section{Introduction} \label{section:introduction}

With the conversion of mainstream media information from text and images into videos, the number of videos on the Internet grows rapidly in recent years. Therefore video analysis evolves into a more important task and attracts much attention from both academy and industry. As a vital area in video analysis, temporal action detection (TAD) aims to localize and recognize action instances in untrimmed long videos. TAD plays an important role in a large number of practical applications, such as video caption~\cite{Dense-Captioning,BAFDVC} and content-based video retrieval~\cite{Frozen-in-time,MMT}.

Recently, a number of methods have been proposed to push forward the state-of-the-art of TAD, which can be mainly divided into three types: anchor-based~\cite{R-C3D,GTAN,PBRNet}, bottom-up~\cite{BSN,BMN,BU-TAL}, and anchor-free~\cite{AFSD,RTD-Net,ActionFormer} methods. 
Although anchor-free methods show stronger competitiveness than others with simple architectures and superior results, they still suffer from the temporal misalignment between the classification and localization tasks.

Current anchor-free frameworks mainly formulate TAD into two tasks: localization and classification. The localization task is designed to generate action proposals, and the classification task is expected to predict action category probabilities which is naturally used as ranking scores in non-maximum suppression (NMS). However, classification and localization tasks usually adopt different training targets. The feature that activates the classification confidence may lack information beneficial to localization, which inevitably leads to misalignment between classification and localization.
To illustrate this phenomenon, we present a case on THUMOS14~\cite{THUMOS14} in Fig.~\ref{fig:motivation_a}, where a proposal with the highest classification score fails to locate the ground truth action. 
This suggests that the classification score can't accurately represent localization quality.
Under this circumstance, accurate proposals may have lower confidence scores and be suppressed by less accurate ones when NMS is conducted. To further demonstrate the importance of accurate score, we replace predicted classification score of action proposals with the actual proposal quality score, which is tIoU between proposal and corresponding ground-truth. As shown in Tab.~\ref{fig:motivation_b}, mAP is greatly improved, which suggests that accurate proposals may not be retrieved due to inaccurate scores.

Recent attempts adopt an additional branch to predict tIoU between proposal and the corresponding ground truth~\cite{AFSD} or focus on the center of an action instance~\cite{ActionFormer}. Although notable improvement is obtained, there is still a huge gap between the performance of previous methods and ideal performance. 
We notice that previous methods mainly rely on the region view which only considers global features of proposals and ignore local appearance and motion evolvement, which increases the difficulty of recognizing boundary location accurateness, especially for actions with long duration.

\begin{table}
  \caption{Oracle experiment results. $G_{\text{score}}$ means proposals scores are replaced by tIoU between proposal and corresponding ground-truth.}
  \label{fig:motivation_b}
  \begin{tabular}{crrrrrr} 
    \toprule
    $G_{\text{score}}$  & 0.3   & 0.4   & 0.5   & 0.6   & 0.7   & Avg. \\
    \midrule
    $\times$            & 60.4  & 54.9  & 46.4  & 35.2  & 21.5  & 43.7 \\
    $\checkmark $       & 93.4  & 92.0  & 88.3  & 82.3  & 72.8  & 85.8\\
    \bottomrule
  \end{tabular}
  \vspace{-0.4cm}
\end{table}

In this paper, we propose a new framework that gives insights into moment and region views simultaneously to align two tasks by estimating reliable proposal quality.
First, we propose Boundary Evaluate Module (BEM) to acquire boundary qualities of proposals from a moment view.
Specifically, BEM focuses on local appearance and motion evolvement for predicting the boundary quality of each temporal location which reflects the distance between the current location to the location of the action boundary.
Then, the quality of the generated proposal is calculated by its boundary qualities.
However, the duration of realistic actions can vary from a few seconds to minutes and the localization quality of short actions is more sensitive to the boundary error than long actions. 
To address this, multi-scale boundary quality is adopted in BEM in a divide-and-conquer way which assigns a suitable scale for each proposal depending on its duration. 
For the region view, we propose a simple but effective module named Region Evaluate Module (REM), which employed the sampled features in proposals as the proposal feature representation and refines proposals.
In particular, REM obtains aligned features by sampling at three locations within the action proposals, which contain more contextual information beneficial to estimate reliable proposal quality and accurate boundary.
The proposed \textbf{B}oundary Evaluate Module and \textbf{R}egion \textbf{E}valuate \textbf{M}odule (BREM) are generic, and they can be integrated with other anchor-free TAD methods to achieve better results.
To validate the effectiveness of BREM, we conduct experiments on two popular mainstream datasets THUMOS14~\cite{THUMOS14} and ActivityNet-1.3~\cite{activitynet}. 
By combining BREM with a basic anchor-free TAD framework proposed by~\cite{A2Net}, we achieve an absolute improvement of $3.6\%$ $m$AP$@$Avg on THUMOS14. When integrating with the state-of-the-art TAD framework ActionFormer~\cite{ActionFormer}, we achieve a new state-of-the-art (63.6\% $m$AP$@$Avg) on THUMOS14 and competitive result (36.2\% $m$AP$@$Avg) on ActivityNet-1.3.

Overall, the contributions of our paper are following:
\textbf{1)} Boundary Evaluate Module (BEM) is proposed to predict multi-scale boundary quality and offer proposal quality from a moment perspective.
\textbf{2)} By introducing Region Evaluate Module (REM), the aligned feature of each proposal are extracted to estimate localization quality in a region view and further refine the locations of action proposals. 
\textbf{3)} The combination of BEM and REM (BREM) makes full use of moment view and region view for estimating reliable proposal quality and it can be easily integrated with other TAD methods with consistent improvement, where a new state-of-the-art result on THUMOS14 and a competitive result on ActivityNet-1.3 are achieved.

\section{Related Work}

\paragraph{Anchor-Based Method.}
Anchor-based methods rely on predefined multiple anchors with different durations and the predictions refined from anchors are used as the final results. 
Inheriting spirits of Faster R-CNN~\cite{FasterRCNN}, R-C3D~\cite{R-C3D} first extracts features at each temporal location, then generates proposals and applies proposal-wise pooling, after that it predicts category scores and relative offsets for each anchor. In order to accommodate varied action durations and enrich temporal context, TAL-Net~\cite{TAL} adopts dilation convolution and scale-expanded RoI pooling.
GTAN~\cite{GTAN} learns a set of Gaussian kernels to model the temporal structure and a weighted pooling is used to extract features.
PBRNet~\cite{PBRNet} progressively refines anchor boundary by three cascaded detection modules: coarse pyramidal detection, refined pyramidal detection, and fine-grained detection.
These methods require predefined anchors which are inflexible because of the extreme variation of action duration.

\paragraph{Bottom-up Method.} Bottom-up methods predict boundary probability for each temporal location, then combines peak start and end to generate proposals. Such as BSN~\cite{BSN}, it predicts start, end, and actionness probabilities and generates proposals, then boundary-sensitive features are constructed to evaluate the confidence of whether a proposal contains an action within its region.
BMN~\cite{BMN} employs an end-to-end framework to generate candidates and confidence scores simultaneously.
BU-TAL~\cite{BU-TAL} explores the potential temporal constraints between start, end, and actionness probabilities.
Some methods, such as~\cite{TCANet,ContextLoc,P-GCN} adopt generated proposals by BSN or BMN as inputs and further refine the boundary and predict more accurate category scores.
Our method is inspired by bottom-up frameworks, but we utilize boundary probability to estimate proposal quality instead of generating proposals.

\paragraph{Anchor-Free Method.} 
Benefiting from the successful application of the anchor-free object detection~\cite{YOLO,FCOS}, anchor-free TAD methods have an increasing interest recently which directly localize action instances without predefined anchors. A2Net~\cite{A2Net} explores the combination of anchor-based and anchor-free methods. 
AFSD~\cite{AFSD} is the first purely anchor-free method that extracts salient boundary features using a boundary pooling operator to refine action proposals and a contrastive learning strategy is designed to learn better boundary features.
Recent efforts aim to use Transformer for TAD. For example, RTD-Net~\cite{RTD-Net} and TadTR~\cite{TadTR} formulate the problem as a set prediction similar to DETR~\cite{DETR}.
ActionFormer~\cite{ActionFormer} adopts a minimalist design and replaces convolution networks in the basic anchor-free framework with Transformer networks.
Our method belongs to anchor-free methods and is easily combined with anchor-free frameworks to boost the performance.

\begin{figure*}[tbp]
    \centering
    \includegraphics[width=\textwidth]{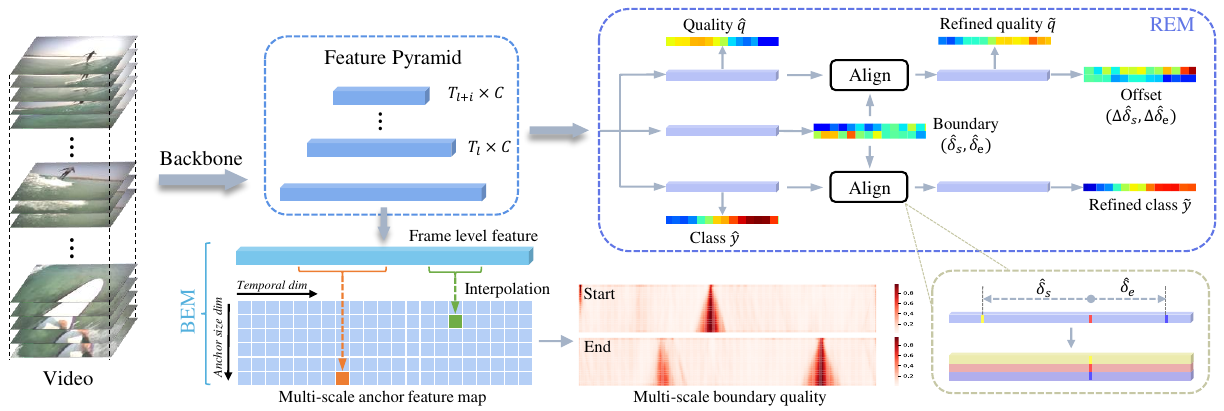}
    \caption{Illustration of the proposed BREM. Untrimmed videos are first fed into the backbone to generate the 1D temporal feature, which is used to construct the feature pyramid and frame-level feature. REM adopts each pyramid feature as input and generates coarse proposals and scores. Then the aligned feature is used for refinement of action location and scores. In parallel, BEM acquires the frame-level feature as input and produces the multi-scale boundary quality map for localization quality prediction.}
    \Description{Waiting to fill.}  
    \label{fig:framework}
    \vspace{-0.3cm}
\end{figure*}

\section{Method}

\paragraph{Problem Formulation.} An untrimmed video can be depicted as a frame sequence $X = \{ x_t \}_{t=1}^T$ with $T$ frames. Action annotations in video $X$ consists of $N_g$ action instances $\Psi_X = \{\psi_n, y_n\}_{n=1}^{N_g} $, where $\psi_n = \left( t_{s}, t_{e} \right)$ are timestamp of start and end of the $n$-th action instance respectively and $y_n$ is the class label. The goal of temporal action detection is to locate boundaries of actions and recognize categories which cover $\Psi_X$ as precisely as possible.

\paragraph{Overview.} For an untrimmed video denoted as $X = \{ x_t \}_{t=1}^T$, a convolution backbone (\textit{e.g.}, I3D~\cite{I3D}, C3D~\cite{C3D}.) is used to extract 1D temporal feature $f \in \mathbb{R}^{T / v \times C}$, where $T$, $C$, $v$ denote video frame, feature channel and stride. Then, up-sample is used to $f$ for acquiring frame level feature $f_{F}$. Multi-scale boundary quality of start and end $\hat{P}_{s/e}$ are predicted by $f_{F}$ (Sec.~\ref{section:BEM}). 
Parallel, several temporal convolutions are used on $f$ to generate the hierarchical feature pyramid. For each hierarchical feature, a shared detection head is applied to predict coarse proposals and category confidences. After that, the aligned feature is extracted for each coarse proposal to refine boundaries and scores (Sec.~\ref{section:REM}). The boundary quality of each proposal is interpolated on $\hat{P}_{s/e}$ according to the temporal location of boundaries.

\subsection{Basic Anchor-free Detector} \label{section:baseline}

Following recent object detection methods~\cite{FCOS} and TAD methods~\cite{A2Net,AFSD}, 
we build a basic anchor-free detector as our baseline, which contains a backbone, a feature pyramid network, and heads for classification and localization.

We adopt I3D network~\cite{I3D} as the backbone since it achieves high performance in action recognition and is widely used in previous action detection methods~\cite{AFSD,BU-TAL}.
The feature output of backbone is denoted as $f \in \mathbb{R}^{T / v \times C}$. Then, $f$ is
used to build hierarchical feature pyramid by applying several temporal convolutions. The hierarchical pyramid features are denoted as $\{ f^l \in \mathbb{R}^{T / v_l \times C}  \}_{l=1}^{L}$, where $l$ means $l$-th layer of feature pyramid and $v_l$ is the stride for the $l$-th layer.

The heads for classification and localization consist of several convolution layers which are shared among each pyramid feature. 
For details, for $l$-th pyramid feature, classification head produces category score $\hat{y} \in \mathbb{R}^{T / v_l \times \mathcal{C}}$, where $\mathcal{C}$ is the number of classes.
And localization head predicts distance between current temporal location to action boundaries, denoted as $\{ ( \hat{\delta}_{s,t}, \hat{\delta}_{e,t}) \}_{t=1}^{T / v_l}$.
Then action detection results are $\{ (c_t, s_t, e_t) \}_{t=1}^{T / v_l}$, where
\begin{equation}
  c_t = \text{arg max}(\hat{y}_t),\ s_t = t - \hat{\delta}_{s,t},\ e_t = t + \hat{\delta}_{e,t}.
  \label{eq:decode_boundary}
\end{equation}
Following AFSD~\cite{AFSD}, the quality branch is also adopted in the baseline model which is expected to suppress low quality proposals.

Based on this baseline model, we further propose two modules named Boundary Evaluate Module (BEM) and Region Evaluate Module (REM) to address the issue of misalignment between classification confidence and localization accuracy. 
Noteworthily, the proposed BEM and REM are generic and easily combined not only with the above baseline framework but also with other anchor-free methods that have a similar pipeline.
The details of BEM and REM would be explained in the rest of this section.

\subsection{Boundary Evaluate Module} \label{section:BEM}

As discussed in Sec.~\ref{section:introduction}, the misalignment between classification confidence and localization accuracy would lead detectors to generate inaccurate detection results. 
To address this, we propose Boundary Evaluate Module (BEM) to extract features and predict action boundary quality maps from a moment view which is complementary to the region view,
thus it can provide more reliable quality scores of proposals.

\paragraph{Single-scale Boundary Quality.} 
Boundary quality maps provide localization quality scores for each temporal location. The quality score is only dependent on the distance from the current location to the location of the action boundary of ground truth. 

\begin{figure}
  \centering
  \includegraphics[width=\columnwidth]{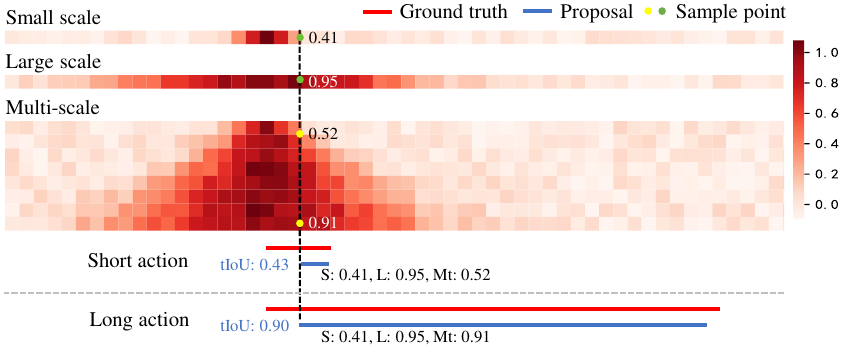}
  \caption{Comparison between single-scale and multi-scale boundary quality maps. For the short proposal and the long proposal, their tIoU are $0.43$ and $0.90$, and their boundary quality scores of small scale, large scale and multi-scale are $(0.41, 0.95, 0.52)$ and $(0.41, 0.95, 0.91)$.}
  \Description{Waiting to fill.}  
  \label{fig:multi-scale}
  \vspace{-0.4cm}
\end{figure}

In practice, we set predefined anchors\footnote{The \textit{anchor} in anchor-based TAD methods is used to describe potential action instances, while we use \textit{anchor} to calculate boundary quality. Thus our method still belongs to the anchor-free method.} at each temporal location, denoted as $\{ a^t \}_{t=1}^T$, where $a^t = [t - r/2, t + r/2]$ denotes the anchor at $t$-th temporal location and $r$ is the predefined anchor size.
For a video with action ground truth $\{ \left( t_{s, n}, t_{e, n} \right) \}_{n=1}^{N_g}$, we define start and end region for $n$-th instance as $g_s^n = [t_{s, n} - r / 2, t_{s, n} + r / 2]$ and $g_e^n = [t_{e, n} - r / 2, t_{e, n} + r / 2]$. The boundary quality maps for start boundary and end boundary $P_s, P_e \in \mathbb{R}^T$ are calculated by
\begin{equation}
  \begin{split}
    P_s^t &= \begin{matrix}
      \max_{n \in N_g} \text{tIoU} (a^t, g_s^n),
    \end{matrix} \\
    P_e^t &= \begin{matrix}
      \max_{n \in N_g} \text{tIoU} (a^t, g_e^n),
    \end{matrix}
  \end{split}
  \label{eq:single_scale_boundary_quality}
\end{equation}
where tIoU is temporal IoU.
The parameter $r$ controls the region size, examples for small and large $r$ are shown in Fig.~\ref{fig:multi-scale} denoted as \textit{Small scale} and \textit{Large scale} separately.
In this way, each score in the quality map indicates the location precision of the start or end boundary. In the inference phase, proposal boundary quality is acquired by interpolation at the corresponding temporal location.

Previous works~\cite{PBRNet,VSGN} formulate the prediction of boundary probability as a binary classification task that can't reflect the relative probability differences between two different locations. However, we define precise boundary quality using tIoU between the predefined anchor and boundary region. Moreover, previous works define positive locations by action length (\textit{e.g.}, locations lie in $[s - d/10, s + d/10]$ are positive samples in \cite{PBRNet} and \cite{VSGN}, where $d$ and $s$ are action length and start location of ground-truth). Thus, the model has to acquire the information of the duration of actions. But it is difficult because of the limited reception field, especially for long actions. So, the definition of boundary quality in Eq.~\ref{eq:single_scale_boundary_quality} is regardless of the duration of actions.
Another weakness of previous works is that they define the action boundary using a small region which leads to that only the proposal boundary closing to the ground-truth boundaries being covered.
In this work, we can adjust $\tau$ to control the region size. We demonstrate that small region size is harmful to performance in our ablation.

\paragraph{Multi-scale Boundary Quality.} 
Actions with different duration require different sensitivity to the boundary changes. 
Fig.~\ref{fig:multi-scale} helps us to illustrate this. 
If we use \textit{Small scale}, a short proposal and a long proposal (blue lines) with the same localization error of start boundary acquire the same boundary qualities of 0.41, but the actual tIoU of the long proposal is 0.9.
Similarly, if we use \textit{Large scale}, these two proposals acquire boundary qualities of 0.95, but the actual tIoU of the short proposal is 0.57. Thus, single-scale boundary quality is suboptimal for varied action duration. The scale should dynamically adapt the duration of actions.
To address this, we expand the single-scale boundary quality maps into quality maps with multi-scale anchors. Thus, for a proposal, we can choose a suitable anchor depending on its duration  (as yellow points show in Fig.~\ref{fig:multi-scale}).

In detail, start and end boundary quality maps are extended to two dimensions corresponding to temporal time steps and anchor scales, denoting as $P_s, P_e \in \mathbb{R}^{T \times I}$, where $I$ is the number of predefined anchors.
We predefine multiple anchors with different size at each temporal location, denoting as $\{ A^t \}_{t=1}^T$, where $A^t = \{a^{t,1}, \cdots, a^{t, I} \}$ denoting $I$ predefined anchors.
The anchor size is defined as
\begin{equation}
    R = \{ r_{min}, r_{max}, I \},
\end{equation}
representing $I$ evenly spaced number from $r_{min}$ to $r_{max}$, where $r_{max}$ and $r_{min}$ indicate the maximum and minimum anchor scale. 
In this paper, $r_{min}$ is set as 1 that corresponds to the interval time between adjacent input video frames and $r_{max}$ depends on the distribution of duration of the actions in datasets.
We conduct ablation studies about the selection of $r_{max}$ in Sec.~\ref{section:ablation}.
Thus the $i$-th anchor at $t$ is $a^{t,i} = [t - r_i/2, t + r_i / 2]$. As for a ground-truth $(t_{s, n}, t_{e, n})$, its start and end region for $i$-th anchor can be denoted as $g_s^{n, i} = [t_{s, n} - r_i / 2, t_{s, n} + r_i / 2]$ and $g_e^{n, i} = [t_{e, n} - r_i / 2, t_{e, n} + r_i / 2]$. Then the multi-scale quality maps $P_s, P_e \in \mathbb{R}^{T \times I}$ are calculated by
\begin{equation}
  \begin{split}
    P_s^{t,i} &= \begin{matrix}
      \max_{n \in N_g} \text{tIoU} (a^{t,i}, g_s^{n,i})
    \end{matrix} \\
    P_e^{t,i} &= \begin{matrix}
      \max_{n \in N_g} \text{tIoU} (a^{t,i}, g_e^{n,i})
    \end{matrix}
  \end{split}
  \label{eq:bem_label}
\end{equation}

In the inference phase, the boundary quality of the proposal is obtained by bilinear interpolation according to the boundaries location and the proposal duration (See Sec.\ref{section:Training_inference}).

\paragraph{Implementation.}
To predict multi-scale boundary quality maps, as shown in Fig.\ref{fig:framework},
the backbone feature is first fed into an up-sampling layer and several convolution layers to get the frame-level feature $f_{F} \in \mathbb{R}^{T \times C}$ with a higher temporal resolution, which is beneficial to predict quality score of the small anchor.
Because the anchor scales may have a large range and different scales need different receptive fields, we adopt a parameter-free and efficient method to generate features.
In detail, we use linear interpolation in each anchor to obtain the multi-scale anchor feature map, denoted as $M \in \mathbb{R}^{T \times I \times N \times C}$.
In particular, for $M(t, i) \in \mathbb{R}^{N \times C}$, we uniformly sample $N$ features in the scope $\left[t - r_i/2, t + r_i/2\right]$ from $f_F$ which ensures that the receptive field matches the anchor size. This procedure of interpolation can be efficiently achieved by matrix product\cite{BMN}.
After the multi-scale anchor feature map $M$ is obtained, we apply max pooling on the $N$ sampled features and a $1 \times 1$ convolution to extract anchor region representation :
\begin{equation}
    M_B = \text{Conv}(\text{MaxPool}(M)),
\end{equation}
where $M_B \in \mathbb{R}^{T \times I \times C}$.
Finally, two boundary score maps are obtained based on $M$ as follows:
\begin{equation}
    \begin{split}
        \hat{P_s} =& \sigma (f_s(M_B)) \\
        \hat{P_e} =& \sigma (f_e(M_B))
    \end{split}
\end{equation}
where $f_s(\cdot)$ and $f_e(\cdot)$ are convolution layers and $\sigma(\cdot)$ is sigmoid function.

\paragraph{Training.} We denote label maps for $\hat{P}_s$ and $\hat{P}_e$ as $O_s, O_e \in \mathbb{R}^{T \times I}$ respectively. 
The label maps is computed by Eq.~\ref{eq:bem_label}.
We take points where $O_{s/e} > 0$ as positive.
L2 loss function is adopted to optimize BEM, which is formulated as follows:
\begin{align}
    \begin{split}
        &\ell_{bem} = 0.5 \cdot (\ell_s + \ell_e), \\
        &\ell_{s/e} = \frac{1}{\left| \mathcal{N}^{s/e} \right|} \sum_{(t, i) \in \mathcal{N}^{s/e}}\left( O_{t, i}^{s/e} - \hat{P}_{t, i}^{s/e} \right)^2 ,
    \end{split}
\end{align}
where $\mathcal{N}^{s/e}$ is the set of positive points.

\begin{table*}[tbp]
  \centering
  \caption{Comparison with state-of-the-art methods on THUMOS14. Average mAP is computed with tIoU thresholds in $[0.3:0.1:0.7]$. The best results are in bold. We integrate BREM with two typical frameworks, baseline (\textit{Base}) (Sec.~\ref{section:baseline}) and ActionFormer~\cite{ActionFormer}. Our method achieves a new state-of-the-art performance on THUMOS14.}
  \label{tab:THUMOS}
  \begin{tabular}{@{}lllrrrrrr@{}}  
  \toprule
  Type & Model                        & Feature         & 0.3   & 0.4   & 0.5  & 0.6  & 0.7  & Avg. \\
  \midrule
  \multirow{6}{*}{Anchor-based}
  & R-C3D~\cite{R-C3D}                & C3D~\cite{C3D}  & 44.8  & 35.6  & 28.9 & -    & -    & -    \\
  & GTAN~\cite{GTAN}                  & P3D~\cite{P3D}  & 57.8  & 47.2  & 38.8 & -    & -    & -    \\
  & PBRNet~\cite{PBRNet}              & I3D~\cite{I3D}  & 58.5  & 54.6  & 51.3 & 41.8 & 29.5 & 47.1 \\
  & A2Net~\cite{A2Net}                & I3D~\cite{I3D}  & 58.6  & 54.1  & 45.5 & 32.5 & 17.2 & 41.6  \\
  & VSGN~\cite{VSGN}                  & TS~\cite{TS}    & 66.7  & 60.4  & 52.4 & 41.0 & 30.4 & 50.2   \\ 
  & G-TAD~\cite{G-TAD}                & TS~\cite{TS}    & 54.5  & 47.6  & 40.2 & 30.8 & 23.4 & 39.3  \\
  \midrule
  \multirow{6}{*}{Bottom-up}
  & BSN~\cite{BSN}                    & TS~\cite{TS}    & 53.5  & 45.0  & 36.9 & 28.4 & 20.0 & 36.8  \\
  & BMN~\cite{BMN}                    & TS~\cite{TS}    & 56.0  & 47.4  & 38.8 & 29.7 & 20.5 & 38.5  \\
  & BC-GNN~\cite{BC-GNN}              & TS~\cite{TS}    & 57.1  & 49.1  & 40.4 & 31.2 & 23.1 & 40.2  \\
  & BU-TAL~\cite{BU-TAL}              & I3D~\cite{I3D}  & 53.9  & 50.7  & 45.4 & 38.0 & 28.5 & 43.3  \\
  & ContextLoc~\cite{ContextLoc}      & I3D~\cite{I3D}  & 68.3  & 63.8  & 54.3 & 41.8 & 26.2 & 50.9  \\
  & TCANet~\cite{TCANet}          & TS~\cite{TS}    & 60.6  & 53.2  & 44.6 & 36.8 & 26.7 & 44.4  \\
  \midrule
  \multirow{8}{*}{Anchor-free}
  & AFSD~\cite{AFSD}                  & I3D~\cite{I3D}  & 67.3  & 62.4  & 55.5 & 43.7 & 31.1 & 52.0   \\
  & RTD-Net~\cite{RTD-Net}            & I3D~\cite{I3D}  & 68.3  & 62.3  & 51.9 & 38.8 & 23.7 & 49.0  \\
  & TadTR~\cite{TadTR}                & I3D~\cite{I3D}  & 62.4  & 57.4  & 49.2 & 37.8 & 26.3 & 46.6 \\
  & ActionFormer~\cite{ActionFormer}  & I3D~\cite{I3D}  & 75.5  & 72.5  & 65.6 & 56.6 & 42.7 & 62.6  \\
  & \textbf{Base}                     & I3D~\cite{I3D}  & 68.5  & 63.7  & 56.6 & 45.8 & 31.0 & 53.1  \\
  & \textbf{Base+BREM}               & I3D~\cite{I3D}  & 70.7  & 66.1  & 60.0 & 50.1 & 36.4 & 56.7  \\
  & \textbf{ActionFormer+BREM}       & I3D~\cite{I3D}  & \textbf{76.5} & \textbf{73.2} & \textbf{66.9} & \textbf{57.7} & \textbf{43.7} & \textbf{63.6} \\
  \bottomrule
  \end{tabular}
  \vspace{-0.2cm}
\end{table*}

\subsection{Region Evaluate Model} \label{section:REM}

\begin{figure}
    \centering
    \includegraphics[width=0.85\columnwidth]{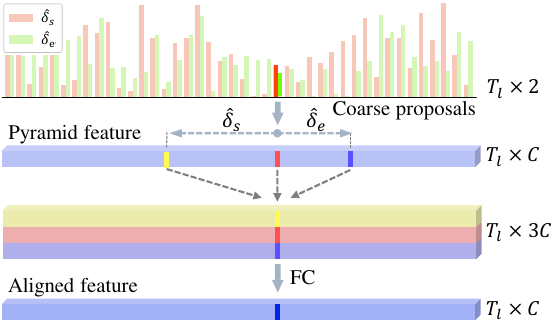}
    \caption{Illustration of feature alignment. According to coarse proposals, sample three features at $\{ t - \hat{\delta}_s, t, t + \hat{\delta}_e \}$ and aggregate them by a fully-connected layer.}
    \Description{Waiting to fill.}  
    \label{fig:align}
    \vspace{-0.3cm}
\end{figure}

BEM estimates the localization quality of proposals in the moment view that focuses more on local appearance and motion evolvement. Although it achieves considerable improvement, as illustrated in Tab.~\ref{tab:components}, we believe that feature of the region view can provide rich context information which is beneficial to the prediction of localization quality. Therefore, we propose Region Evaluate Module (REM), as shown in the right part of Fig.~\ref{fig:framework}, which first predicts coarse action proposals and then extracts features of proposals to predict localization quality scores, action categories, and boundary offsets.

Specifically, REM predicts coarse action offset $(\hat{\delta}_s, \hat{\delta}_e)$, action categories $\hat{y}$ and quality score $\hat{q}$ for each temporal location (omitting subscript standing for temporal location for simplicity). For a location $t$ with coarse offset prediction which indicates the distance to start and end of the action boundaries, the corresponding proposal can be denoted as $\hat{\psi}=(t - \hat{\delta}_s, t + \hat{\delta}_e)$.
Then three features are sampled from pyramid feature at $\{ t - \hat{\delta}_s, t, t + \hat{\delta}_e \}$ via linear interpolation and aggregated by a fully-connected layer. This procedure is illustrated in Fig.~\ref{fig:align}. Based on the aggregated feature, BEM produces refined boundary offsets $(\Delta \hat{\delta}_s, \Delta \hat{\delta}_e)$, quality scores $\tilde{q}$ and category scores $\tilde{y}$.
The final outputs can be obtained by
\begin{equation}
    \begin{split}
        y &= 0.5 \cdot (\hat{y} + \tilde{y}),\\
        q &= 0.5 \cdot (\hat{q} + \tilde{q}), \\
        \psi &= (t-\hat{\delta}_s - 0.5 \cdot \Delta \hat{\delta}_s  \hat{w}, t+\hat{\delta}_e + 0.5 \cdot \Delta \hat{\delta}_e \hat{w})
    \end{split}
    \label{eq:rem_out}
\end{equation}
where $\psi$, $y$, $q$ are final action proposal, action category score and location quality score respectively and $\hat{w} = \hat{\delta}_s + \hat{\delta}_e$.

\paragraph{Training.}
The loss of REM is formulated as:
\begin{equation}
    \ell_{rem} = \hat{\ell}_{loc} + \lambda \hat{\ell}_{cls} + \gamma \hat{\ell}_{q} + \tilde{\ell}_{loc} + \lambda \tilde{\ell}_{cls} + \gamma \tilde{\ell}_{q},
\end{equation}
where $\lambda$, $\gamma$ are loss weight.
$\hat{\ell}_{cls}$ and $\tilde{\ell}_{cls}$ are focal loss~\cite{FocalLoss} for category prediction. 
$\hat{\ell}_{q}$ and $\tilde{\ell}_{q}$ are loss of quality prediction, which is implemented by binary cross entropy loss. tIoU between proposal and corresponding ground-truth is adopted as target of quality prediction:
\begin{equation}
  \hat{\ell}_q = \frac{1}{ \left| \mathcal{N}_{pos} \right|} \begin{matrix} \sum_{t \in \mathcal{N}_{pos}} \text{BCE}(\hat{q}_t, \text{tIoU}(\psi_t, \hat{\psi}_t))  \end{matrix},
\end{equation}
where $\psi_t$ is ground-truth for location $t$.
$\hat{\ell}_{loc}$ is generalized IoU loss~\cite{GIoULoss} for location prediction of initial proposal and $\tilde{\ell}_{loc}$ is L1 loss for offset prediction of the refining stage:
\begin{equation}
  \begin{split}
    \hat{\ell}_{loc} &= \frac{1}{ \left| \mathcal{N}_{pos} \right|} \begin{matrix} \sum_{t \in \mathcal{N}_{pos}} ( 1 - \text{GIoU}(\psi_t, \hat{\psi}_t)) \end{matrix}, \\
    \tilde{\ell}_{loc} &= \frac{1}{ \left| \mathcal{N}_{pos} \right|} \begin{matrix} \sum_{t \in \mathcal{N}_{pos}} (| \Delta \hat{\delta} -  \Delta \delta |) \end{matrix}
  \end{split}
\end{equation}
where $\mathcal{N}_{pos}$ indicates the ground-truth action locations, and $\Delta \delta = 2 \cdot (\delta -  \hat{\delta}) / \hat{w}$, $\hat{w}$ is coarse proposal length.

\subsection{Training and Inference} \label{section:Training_inference}

\paragraph{Training details.} Since there are mainly two different strategies for video feature extraction, including online feature extraction~\cite{R-C3D,AFSD} and offline feature extraction~\cite{BSN,BMN,TCANet}, we adopt different training methods for them. For frameworks using the online feature extractor, BEM and REM are trained jointly with the feature extractor in an end-to-end way. 
The total train loss function is
\begin{equation}
  \ell = \ell_{rem} + \eta \ell_{bem},
\end{equation}
where $\eta$ is used to balance loss.
As for methods with the offline feature extractor, since BEM is independent of other branches, we individually train BEM and other branches, then combine them in the inference phase for better performance.

\paragraph{Inference.} The final outputs of REM is calculated by Eq.~\ref{eq:rem_out}. Thus, the generated proposals can be denoted as $\{ (y, q, \psi)_n \}_{n=1}^N$, where $\psi = (t_s, t_e)$ and $N$ is the number of proposals.
In order to obtain boundary quality, we define a function that generates index of appropriate anchor scale in multi-scale boundary quality map according to the action duration, denoted as $f(d)$. We adopt a simple linear mapping:
\begin{equation}
  \begin{split}
    f(d) &= \frac{r - r_i}{r_{i+1} - r_i} + i, \\
    r &= d / \tau, \\
    s.t. \ \ r_i &\leqslant r \leqslant r_{i+1},
  \end{split}
  \label{eq:bem_inference}
\end{equation}
where $\tau$ is a predefined mapping coefficient.
For a proposal, $\tau$ controls the anchor size used by it. We explore the influence of $\tau$ in our ablation.
Then start and end boundary quality are acquired by bilinear interpolation,
\begin{equation}
  p_{s,d} = \textit{Intep}(P_s, (t_s, f(d))), \ p_{e,d} = \textit{Intep}(P_e, (t_e, f(d))),
\end{equation}
where $Intep$ is bilinear interpolation and $d=t_e - t_s$ is the length of proposal.
After fusing these scores, the final proposals is denoted as $\{(y \cdot q \cdot \sqrt{p_{s, d} \cdot p_{e, d}}, \psi) \}_{n=1}^N$.

\section{Experiments}
\paragraph{Dataset.} The experiments are conducted on two popularly used datasets, THUMOS14~\cite{THUMOS14} and ActivityNet-1.3~\cite{activitynet}. THUMOS14 contains 200 untrimmed videos in the validation set and 212 untrimmed videos in the testing set with 20 categories. 
Following previous works~\cite{BMN,BSN,BU-TAL}, we train our models on the validation set and report the results on the testing set.
ActivityNet-1.3 contains 19,994 videos of 200 classes with about 850 video hours. The dataset is split into three subsets, about 50$\%$ for training, and 25$\%$ for validation and testing.
Following~\cite{BSN,BMN,G-TAD}, the training set is used to train the models, and results are reported on the validate set.

\paragraph{Implementation Details.} For THUMOS14 dataset, we sample 10 frames per second (fps) and resize the spatial size to 96 $\times$ 96. Same as the previous works~\cite{BMN,AFSD}, sliding windows are used to generate video clips. Since nearly $98\%$ action instances are less than 25.6 seconds in the dataset, the windows size is set to 256. The sliding windows have a stride of 30 frames in training and 128 frames in testing. 
The feature extractor is I3D~\cite{I3D} pre-trained in Kinetics.
The mean Average Precision (mAP) is used to evaluate performance. The tIoU thresholds of $\left[0.3:0.1:0.7\right]$ are considered for mAP and average mAP. If not noted specifically, we use Adam as optimizer with the weight decay of $10^{-3}$. The batch size is set to 8 and the learning rate is $8\times 10^{-4}$.
As for loss weight, $\eta$, $\lambda$, $\gamma$ are set to 5, 1 and 0.5. 
The anchor scale $R$ and mapping coefficient $\tau$ in BEM are $\{ 1, 50, 20 \}$ and 2.
In the testing phase, the outputs of RGB and Flow are averaged.
The tIoU threshold of Soft-NMS is set as 0.5.

On ActivityNet-1.3, each video is encoded to 768 frames in temporal length and resized to 96 $\times$ 96 spatial resolution. I3D backbone is pre-trained in Kinetics. 
mAP with tIoU thresholds $\left\{ 0.5, 0.75, 0.95 \right\}$ and average mAP with tIoU thresholds $\left[0.5:0.05:0.95\right]$ are adopted. Optimizer is Adam with weight decay of $10^{-4}$. Batch size is 1 and learning rate is $10^{-5}$ for feature extractor and $10^{-4}$ for other components. As for loss weight, $\eta$, $\lambda$, $\gamma$ are set to 5, 1 and 1 repestively. 
The anchor scale $R$ and mapping coefficient $\tau$ in BEM are $\{ 1, 130, 22 \}$ and 2. The tIoU threshold of Soft-NMS is set to 0.85.

In order to validate the generalizability of our method, we also evaluate the performance when integrating BREM with methods using the offline feature extractor. ActionFormer~\cite{ActionFormer} is the latest anchor-free TAD method that shows strong performance. 
Thus we integrate BREM with ActionFormer to validate the effectiveness of BREM. The implementation details are shown in our supplement.

\begin{table}[tb]
    \centering
    \caption{Comparison with state-of-the-art methods on ActivityNet-1.3. Average mAP is computed with tIoU thresholds in $[0.3:0.1:0.7]$. We integrate BREM with two typical frameworks, baseline (Sec.~\ref{section:baseline}) (\textit{Base}) and ActionFormer~\cite{ActionFormer} (\textit{AF}).}
    \label{tab:ActivityNet}
    \begin{tabular}{llrrrr}
        \toprule
        Model  &  Feature & 0.5 & 0.75 & 0.95 & Avg. \\
        \midrule
        \textit{Anchor-based} \\
        \midrule
        R-C3D~\cite{R-C3D}            & C3D~\cite{C3D}  & 26.8 & -    & -    & -     \\
        GTAN~\cite{GTAN}              & P3D~\cite{P3D}  & 52.6 & 34.1 & 8.9  & 34.3  \\
        PBRNet~\cite{PBRNet}          & I3D~\cite{I3D}  & 54.0 & 35.0 & 9.0  & 35.0  \\
        A2Net~\cite{A2Net}            & I3D~\cite{I3D}  & 43.6 & 28.7 & 3.7  & 27.8    \\
        VSGN~\cite{VSGN}              & TS~\cite{TS}    & 52.4 & 36.0 & 8.4  & 35.1  \\
        G-TAD~\cite{G-TAD}            & TS~\cite{TS}  & 50.4 & 34.6 & 9.0  & 34.1  \\
        \midrule
        \textit{Bottom-up} \\
        \midrule
        BSN~\cite{BSN}                & TS~\cite{TS}  & 46.5 & 30.0 & 8.0  & 30.0    \\
        BMN~\cite{BMN}                & TS~\cite{TS}  & 50.1 & 34.8 & 8.3  & 33.9    \\
        BC-GNN~\cite{BC-GNN}          & TS~\cite{TS}  & 50.6 & 34.8 & \textbf{9.4}  & 34.3   \\
        BU-TAL~\cite{BU-TAL}          & I3D~\cite{I3D}  & 43.5 & 33.9 & 9.2  & 30.1    \\
        ContextLoc~\cite{ContextLoc}  & I3D~\cite{I3D} & \textbf{56.0} & 35.2 & 3.6 & 34.2 \\
        TCANet~\cite{TCANet}          & SlowFast~\cite{SlowFast} & 54.3 & \textbf{39.1} & 8.4 & \textbf{37.6} \\
        \midrule
        \textit{Anchor-free} \\
        \midrule
        AFSD~\cite{AFSD}              & I3D~\cite{I3D}  & 52.4 & 35.3 & 6.5  & 34.4     \\
        RTD-Net~\cite{RTD-Net}        & I3D~\cite{I3D}  & 47.2 & 30.7 & 8.6  & 30.8   \\
        TadTR~\cite{TadTR}            & I3D~\cite{I3D}  & 49.1 & 32.6 & 8.5  & 32.3  \\
        ActionFormer (TSP)            & R(2+1)D~\cite{R(2+1)D}  & 54.1 & 36.3 & 7.7  & 36.0 \\
        \textbf{Base}                & I3D~\cite{I3D}  & 52.4 & 34.3 & 5.6  & 33.6\\
        \textbf{Base+BREM}           & I3D~\cite{I3D}  & 52.2 & 35.4 & 5.1  & 34.3   \\
        \textbf{AF+BREM (TSP)}       & R(2+1)D~\cite{R(2+1)D} & 53.7 & 37.9 & 6.9 & 36.2 \\
        \bottomrule
    \end{tabular}
    \vspace{-0.2cm}
\end{table}

\subsection{Main Result}

In this subsection, we compare our models with state-of-the-art methods, including anchor-based (\textit{e.g.}, R-C3D~\cite{R-C3D}, PBRNet~\cite{PBRNet}, VSGN~\cite{VSGN}), bottom-up (\textit{e.g.} BMN~\cite{BMN}, TCANet~\cite{TCANet}), and anchor-free (\textit{e.g.}, AFSD~\cite{AFSD}, RTD-Net~\cite{RTD-Net}) methods. And the features used by these methods are also reported for a more fair comparison, including C3D~\cite{C3D}, P3D~\cite{P3D}, TS~\cite{TS}, I3D~\cite{I3D}, and R(2+1)D~\cite{R(2+1)D}. 

The results on the testing set of THUMOS14 are shown in Tab.~\ref{tab:THUMOS}.
Our baseline achieves 53.1$\%$ $m$AP$@$Avg outperforming most of the previous methods. Based on the strong baseline, BREM absolutely improves 3.6$\%$ from 53.1$\%$ to 56.7$\%$ on $m$AP$@$Avg. 
It can be seen that the proposed BREM acquires improvement on each tIoU threshold compared with the baseline.
Especially on high tIoU thresholds, BREM achieves an improvement of 5.4$\%$ on $m$AP$@0.7$.
Similarly, integrating BREM with ActionFormer~\cite{ActionFormer} provides a performance gain of 1.3$\%$ on $m$AP$@0.5$ and yields a new state-of-the-art performance of 63.6$\%$ on $m$AP$@$Avg.

The results on ActivityNet-1.3 validation set are shown in Tab.~\ref{tab:ActivityNet}.
Integrating BREM with baseline (\textit{Base}) reaches an average $m$AP of 34.3$\%$, which is 0.7$\%$ higher than baseline.
And BREM achieves an average $m$AP of 36.2$\%$ when combined with ActionFormer (\textit{AF}) using the pre-training method from TSP~\cite{TSP}, which is the best result using the features from \cite{R(2+1)D}.
It is worthy to note that BREM brings considerable improvement on middle tIoU thresholds, outperforming ActionFormer by 1.6$\%$ on $m$AP$@0.75$.
TCANet~\cite{TCANet} is the only model better than ours, but it uses the stronger SlowFast feature~\cite{SlowFast} and refines proposals generated by a strong proposal generation method~\cite{BMN}.

\subsection{Ablation Study} \label{section:ablation}

We conduct ablation experiments on THUMOS14 for the RGB model based on the baseline to validate the effectiveness of our method. 
The $m$AP at tIoU=0.5, 0.6 and 0.7, and average $m$AP in $[0.3:0.1:0.7]$ are reported.
Each experiment is repeated three times and the average result is presented to obtain more convincing results.

\begin{table}
  \centering
  \caption{Effectiveness of BEM and REM. The first row represents the result of the baseline model described in Sec.~\ref{section:baseline}.}
  \label{tab:components}
  \begin{tabular}{cc|rrrr}
      \toprule
      BEM         & REM           & 0.5   & 0.6   & 0.7   & Avg. \\
      \midrule
                  &               & 47.0  & 35.4  & 22.9  & 44.2  \\
      $\checkmark$&               & 48.9  & 38.5  & 27.1  & 46.4  \\
                  & $\checkmark$  & 47.4  & 37.4  & 25.0  & 45.4  \\
      $\checkmark$& $\checkmark$  & \textbf{50.2}& \textbf{40.8}& \textbf{29.0}& \textbf{48.3}  \\
      \bottomrule
  \end{tabular}
  \vspace{-0.2cm}
\end{table}

\begin{table}
  \centering
  \caption{The effectiveness of boundary quality. $\{r_{min}, r_{max}, I \}$ represent $I$ evenly spaced numbers from $r_{min}$ to $r_{max}$. The first row indicates the model without boundary quality.}
  \label{tab:multi-scale}
  \begin{tabular}{lc|rrrr}
      \toprule
      Type & Anchor size   & 0.5   & 0.6   & 0.7   & Avg. \\
      \midrule
      w/o             & -     & 47.0  & 35.4  & 22.9  & 44.2  \\
      \midrule
      \multirow{4}{*}{Single-scale}
      & 4             & 45.2  & 34.3  & 21.9  & 42.6 \\
      & 16            & 47.3  & 37.4  & 25.9  & 45.2  \\
      & 28            & \textbf{47.8}  & \textbf{37.5}  & \textbf{26.4}  & \textbf{45.4} \\
      & 40            & 47.5  & 37.4  & 25.7  & 45.1 \\
      \midrule
      \multirow{5}{*}{Multi-scale}
      & $\{ 1, 10, 20 \}$   & 47.0  & 37.1  & 25.5  & 45.0  \\
      & $\{ 1, 20, 20 \}$   & 47.2  & 37.6  & 27.0  & 45.5  \\
      & $\{ 1, 40, 20 \}$   & 48.1  & \textbf{38.9}  & \textbf{27.4}  & 46.3  \\
      & $\{ 1, 50, 20 \}$   & \textbf{48.9}  & 38.5  & 27.1  & \textbf{46.4}  \\
      & $\{ 1, 60, 20 \}$   & 48.6  & 38.6  & 27.2  & \textbf{46.4} \\
      \bottomrule
  \end{tabular}
  \vspace{-0.3cm}
\end{table}

\paragraph{Effectiveness of Model Components} In order to analyze the effectiveness of the proposed BEM and REM, each component is applied in the baseline model gradually. Meanwhile, the result of the combination of BEM and REM is also presented to demonstrate they are complementary to each other. All results are shown in Tab.~\ref{tab:components}. Obviously,
BEM boosts the average $m$AP by 2.2$\%$. The significant improvement brought by BEM confirms that BEM helps to preserve better action proposals based on the more accurate quality score of boundary localization.
Meanwhile, REM improves the average $m$AP by 1.2$\%$. This suggests that aligned features are beneficial for refining more accurate boundaries, classification, and quality scores.
By combining BEM and REM, the performance is further improved from 44.2$\%$ to 48.3$\%$ on $m$AP$@$Avg.
The great complementary result shows that the moment view of BEM and region view of REM are both essential.

\paragraph{Effectiveness of Boundary Quality}
In order to demonstrate the effectiveness of boundary quality, we first analyze its importance by introducing single-scale boundary quality. Then the comparison between single-scale and multi-scale boundary quality is conducted to validate the necessity of introducing more anchor scales. Finally, different settings of boundary anchors are explored.
Results are shown in Tab.~\ref{tab:multi-scale}.
For single-scale boundary quality with anchor size=4, the $m$AP$@$Avg drops from 44.2$\%$ to 42.6$\%$. We conjecture that the reason is that the estimated boundary quality at the most temporal locations can not reflect the actual location quality because of the small anchor size (see Fig.~\ref{fig:multi-scale} \textit{Small scale}).
Increasing the anchor size boosts the performance. The best result is reached with anchor size=28, and further increasing the anchor size harms the performance.
For multi-scale boundary quality, we gradually increase the largest anchor size ($r_{max}$). As shown in Tab.~\ref{tab:multi-scale}, increasing $r_{max}$ improves the performance, and saturation is reached when $r_{max}=50$ because there are few long actions in the dataset thus too large anchors are rarely used.
The above results suggest that our single-scale boundary quality can help preserve better predictions in NMS, but a suitable anchor size has to be carefully chosen.
Contrary to single-scale boundary quality, multi-scale boundary quality introduces further improvement by dividing actions into different appropriate anchor scales depending on their duration. It can be seen that the anchor size of $\{ 1, 50, 20 \}$ brings a 1\% improvement compared with single-scale boundary quality. Furthermore, it is less sensitive to the choice of anchor size. 

\begin{table}
  \centering
  \caption{Effectiveness of each component of REM.}
  \label{tab:REM_ablation}
  \begin{tabular}{l|rrrr}
      \toprule
      Model               & 0.5   & 0.6   & 0.7   & Avg. \\
      \midrule
      REM                 & 47.4  & \textbf{37.4}  & \textbf{25.0}  & \textbf{45.4}  \\
      w/o offset          & 47.2  & 36.7  & 23.5  & 44.6  \\
      w/o quality         & \textbf{47.7}  & 37.1  & 24.7  & 45.3  \\
      w/o classification  & 47.0  & 36.9  & 24.7  & 44.9  \\
      \bottomrule
  \end{tabular}
  \vspace{-0.3cm}
\end{table}

\paragraph{Effectiveness of REM}
Based on the aligned feature, REM refines the location, category score, and localization quality score of each action proposal. We gradually remove each component to show its effectiveness.  The results are shown in Tab.~\ref{tab:REM_ablation}.
Removing offset, quality, and classification drop the performance by 0.8$\%$, 0.1$\%$, and 0.5$\%$ respectively. 
Refinement of location and category score bring more noticeable improvement to the model than quality score.
We preserve quality score refinement in our final model since it can stable the performance and only increases negligible computation.
Previous work~\cite{AFSD} extracts salient boundary feature by boundary max pooling, while we extract the region feature of the proposal by interpolation which is more efficient and shows competitive performance.

\begin{table}
  \centering
  \caption{Ablation study on regional feature extraction method in REM.}
  \label{tab:feat_reduction}
  \begin{tabular}{l|rrrr}
    \toprule
    method      & 0.5   & 0.6   & 0.7   & Avg. \\
    \midrule
    FC          & 47.7  & 37.6  & 26.3  & 45.5  \\
    Mean        & 48.2  & 38.1  & \textbf{27.4}  & 46.1  \\
    Max         & \textbf{48.9}  & \textbf{38.5}  & 27.1  & \textbf{46.4}  \\
    Mean$\&$Max & 48.3  & 38.1  & 26.7  & 46.1  \\
    \bottomrule
  \end{tabular}
  \vspace{-0.2cm}
\end{table}

\begin{table}
  \centering
  \caption{Ablation study on mapping coefficient $\tau$ in BEM.}
  \label{tab:ablation_of_tau}
  \begin{tabular}{l|rrrr}
    \toprule
    $\tau$  & 0.5   & 0.6   & 0.7   & Avg. \\
    \midrule
    0.5     & 48.3  & 37.3  & 25.0  & 45.5 \\
    1.0     & \textbf{49.0}  & 38.1  & 26.0  & \textbf{46.4} \\
    2.0     & 48.9  & \textbf{38.5}  & \textbf{27.1}  & \textbf{46.4} \\
    4.0     & 47.1  & 37.2  & 25.6  & 45.0 \\
    \bottomrule
  \end{tabular}
  \vspace{-0.3cm}
\end{table}

\paragraph{Ablation study on regional feature extraction method in REM}
We explore different feature extraction methods in REM, 
1) FC: all sampled features in each anchor region are concatenated and a fully connected layer is applied to convert them to the target dimension.
2) Mean: the mean operation is applied to all sampled features.
3) Max: the mean operation in Mean is replaced with max.
4) Mean$\&$Max: Mean feature and Max feature are concatenated and a fully connected layer is applied to convert the dimension of the feature.
The results are shown in Tab.~\ref{tab:feat_reduction}.
FC is commonly used in previous works~\cite{BMN,BSN++}, but reaches the lowest performance in our experiments. 
Max acquires the best performance of average $m$AP, showing 0.9$\%$, 0.3$\%$ and 0.3$\%$ advantage against FC, Mean and Mean$\&$Max respectively.

\paragraph{Ablation study on mapping coefficient $\tau$ in BEM}
The mapping coefficient $\tau$ in BEM controls the corresponding anchor size of the proposal in the inference phase (See Eq.~\ref{eq:bem_inference}). For a proposal, it will use a smaller scale anchor if enlarging $\tau$. 
We vary the mapping coefficient $\tau \in \{ 0.5, 1.0, 2.0, 3.0\}$ in the inference phase and report the results in Tab.~\ref{tab:ablation_of_tau}.
The performance is stable if $\tau$ equals to 1.0 or 2.0.
Smaller and larger $\tau$ will decrease the performance since the anchor size and the duration of action are not appropriately matched, which also confirms the importance of multi-scale boundary quality.

\section{Conclusion}
In this paper, we reveal the issue of misalignment between localization accuracy and classification score of current TAD methods. To address this, we propose \textbf{B}oundary Evaluate Module and \textbf{R}egion \textbf{E}valuate \textbf{M}odule (BREM), which is generic and plug-and-play. In particular, BEM estimates the more reliable proposal quality score by predicting multi-scale boundary quality in a moment perspective. Meanwhile, REM samples region features in action proposals to further refine the action location and quality score in a region perspective.
Extensive experiments are conducted on two challenging datasets.
Benefiting from the great complementarity of moment and region perspective, BREM achieves state-of-the-art results on THUMOS14 and competitive results on ActivityNet-1.3.

\begin{acks}
This work was supported in part by Next Generation AI Project of China No.2018AAA0100602, in part to Dr. Liansheng Zhuang by National Natural Science Foundation of China (NSFC) under contract No.U20B2070 and No.61976199, and in part to Dr. Houqiang Li by NSFC under contract No.61836011.
\end{acks}

\newpage
\bibliographystyle{ACM-Reference-Format}
\bibliography{references}

\appendix

\section{Experimental detail}

The proposed \textbf{B}oundary Evaluate Module and \textbf{R}egion \textbf{E}valuate \textbf{M}odule (BREM) is generic, and it is easily combined with other anchor-free frameworks to achieve better results. Fig. \ref{fig:framework} illustrates the overall architecture of anchor free methods combined with BREM, where BEM is integrated as an extra component and REM is adopted to replace the original detection head.
In our experiments, BREM is combined with two typical frameworks, a basic anchor-free framework (denoted as \textit{Base}) and \text{ActionFormer}~\cite{ActionFormer}. The implementation details are described in this section.

\subsection{The architecture of Boundary Evaluate Module}
The implementation of Boundary Evaluate Module (BEM) is shown in Tab~\ref{tab:BEM_implementation}. The input feature of BEM is frame level feature $f_{F} \in \mathbb{R}^{C  \times T}$ with the time resolution same as the input of backbone, which preserves more detail information of appearance and motion. In the experiments, the number of sample points $N$ is set to 14 for the balance of efficiency and effectiveness.

\subsection{The architecture of Region Evaluate Module}
As Figure 2 shown in our main paper, the inputs of REM are feature pyramid denoted as $\{ f^l \in \mathbb{R}^{T / v_l \times C}  \}_{l=1}^{L}$. For simplicity, we omit the superscript standing for pyramid layers in the following. For a feature of pyramid (denoted as $f$), BREM predicts coarse action offset $\hat{\delta}$, action categories $\hat{y}$ and quality score $\hat{q}$ by the equations
\begin{equation}
    \begin{split}
        \hat{\delta} &=h_o( g_o (f)), \\
        \hat{y} &=\sigma( h_c( g_c (f))), \\
        \hat{q} &=\sigma( h_q( g_o (f))),
    \end{split}
\end{equation}
where $g_o(\cdot), g_c(\cdot)$ are hidden convolution layer, and $h_o(\cdot), h_c(\cdot), h_q(\cdot)$ are  convolution layer with output channels of $2$ (coarse offset to start and end of action), $\mathcal{C}$ (the number of categories) and $1$ respectively. The hidden layer of $\hat{q}$ is shared with offset prediction branch which is usually adopted in previous work [1]. As for refined boundary offsets $\Delta \hat{\delta}$, quality scores $\tilde{q}$ and category scores $\tilde{y}$, they use a similar method as the above description except that the input feature is  the aligned feature by "Align" module (see Figure 4 in the main paper). This is formulated as
\begin{equation}
    \begin{split}
        f' = \text{Align}(f),& \
        \Delta \hat{\delta} =h_o'( g_o' (f')), \\
        \tilde{y} =\sigma( h_c'( g_c' (f'))),& \
        \tilde{q} =\sigma( h_q'( g_o' (f'))).
    \end{split}
\end{equation}
Finally, the refined action prediction is obtained via Eq. 8 in the main paper.

\begin{figure}[tbp]
  \centering
  \subfigure[Framework of anchor-free methods.]{
    \begin{minipage}{0.75\linewidth}
      \centering
      \includegraphics[width=\linewidth]{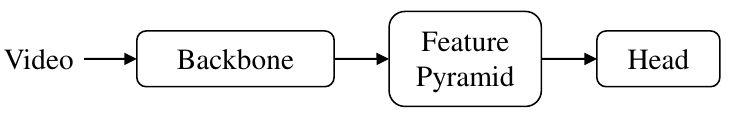}
      \label{fig:framework_base}
    \end{minipage}
  }
  \subfigure[Framework of combination BREM with anchor-free methods.]{
    \begin{minipage}{0.8\linewidth}
      \centering
      \includegraphics[width=\linewidth]{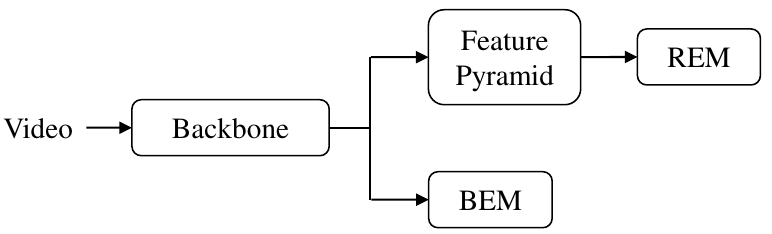}
    \end{minipage}
    \label{fig:framework_brem}
  }
  \Description{Waiting to fill.}  
  \caption{The framework of anchor-free methods and BREM.}
  \label{fig:framework_supp}
\end{figure}

\begin{table}
  \caption{The implementation of Boundary Evaluate Module. $T$, $I$, and $N$ are length of input feature, number of anchors, and number of sample points. $Intep$ denotes sampling features within each anchor.}
  \label{tab:BEM_implementation}
  \begin{tabular}{l|ccc|l}
    \toprule
    layer & kernel & output dim & act & output size \\
    \midrule
    $conv1d_1$ & 3 & 256 & $relu$ & $256 \times T$ \\
    $Intep$    &    &     &       & $256 \times N \times T \times I$ \\
    $MaxPool$  &    &     &       & $256 \times 1 \times T \times I$ \\
    $Squeeze$  &    &     &       & $256 \times T \times I$ \\
    $conv2d_1$ & 1  & 128 & $relu$& $128 \times T \times I$ \\
    $conv2d_2$ & 3  & 128 & $relu$& $128 \times T \times I$ \\
    $conv2d_3$ & 1  & 2 & $sigmoid$& $2 \times T \times I$ \\
    \bottomrule
  \end{tabular}
\end{table}

\subsection{Combination with \textit{Base}}

\begin{figure}
  \centering
  \includegraphics[width=0.9\linewidth]{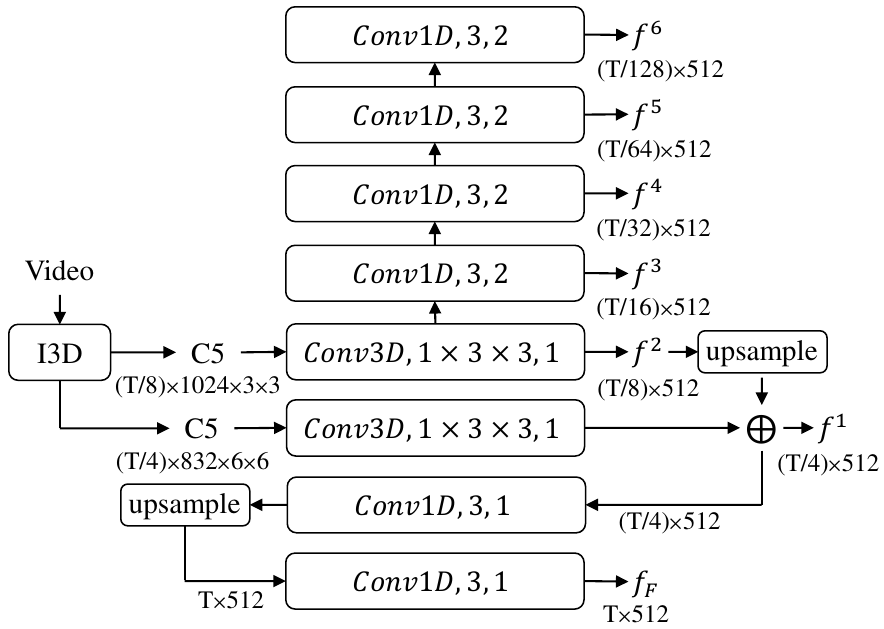}
 \caption{The architecture of feature encoder of \textit{Base}. The convolutional layer is denoted as $Conv, kernel\ size, stride$, and the channel of all convolutional layers is $512$. $\oplus$ denotes element-wise summation.}
  \Description{Null}
  \label{fig:FPN}
\end{figure}

\paragraph{Detailed architecture}
We adopt a previous successful feature pyramid proposed by AFSD~\cite{AFSD} and its architecture is shown in Fig.~\ref{fig:FPN}. I3D~\cite{I3D} is used to extract the semantic feature of videos and the feature of 4$th$ and 5$th$ stage (C4 and C5) are used to generate a feature pyramid. Meanwhile, an up-sample layer and a convolutional layer are used to produce the frame-level feature. Finally, pyramid features and frame level feature are fed into REM and BEM, respectively.

\subsection{Combination with ActionFormer}

\paragraph{Detailed architecture}
Contrary to \textit{Base}, ActionFormer~\cite{ActionFormer} uses off-the-shelf features. Frame level feature $f_F$ is generated by applying a $Conv1D$ with kernel size=3 and stride=1 on pre-encoded video features. Then $f_F$ is fed into BEM and the original head of ActionFormer is replaced with REM.

\paragraph{Implementation}
Since the video feature $f$ is pre-encoded, REM and BEN are separately trained for better performance and convergence. Other training details are same as ActionFormer~\cite{ActionFormer}.

\section{Additional Experiment Results}

\begin{table}
  \caption{Results of different mapping functions.}
  \label{tab:multi_ablation}
  \begin{tabular}{l|rrrr}
    \toprule
    Mapping & 0.5 & 0.6 & 0.7 & Avg. \\
    \midrule
    w/o BEM & 47.0  & 35.4  & 22.9  & 44.2 \\
    $f(d | \tau=2)$ & \textbf{48.9}  & \textbf{38.5}  & \textbf{27.1}  & \textbf{46.4}  \\
    \midrule
    $g(d|D=6)$ & 46.0 & 36.0 & 23.7 & 44.0 \\
    $g(d|D=18)$ & 47.4 & 37.2 & 25.3 & 45.4 \\
    $g(d|D=31)$ & 48.3 & \textbf{37.8} & \textbf{25.9} & \textbf{45.8} \\
    $g(d|D=43)$ & \textbf{48.4} & 37.5 & 25.3 & 45.6 \\
    \bottomrule
  \end{tabular}
\end{table}


\subsection{Additional Ablation Study of Multi-scale Boundary Quality}

In this section, we conduct additional ablation studies to explore the effectiveness of multi-scale boundary quality compared with single-scale boundary quality.

The proposed method uses a linear mapping function $f(d)$ to generate the index of appropriate anchor scale in the multi-scale boundary quality map according to the action duration.
In this section, we explore a special mapping function, denoted as $g(d)$, 
\begin{equation}
  \begin{split}
    g(d) &= \frac{r - r_i}{r_{i+1} - r_i} + i, \\
    r &= D, \\
    s.t. \ \ r_i &\leqslant r \leqslant r_{i+1}, 
    r_{min} \leqslant D \leqslant r_{max}, \\
    \label{eq:g_d}
  \end{split}
\end{equation}
where $D$ is a hyper-parameter which indicates the anchor size used in the inference phase. This mean that all proposals use the same anchor size $D$. Unlike $f(d)$ that assigns proposals to anchors with appropriate size, $g(d)$ assigns all proposals to a same anchor. The comparison between $f(d)$ and $g(d)$ can demonstrate the effectiveness of assigning proposals of different duration to appropriate anchors. In the experiments on THUMOS14, we set 20 anchors from 1 to 50 with even interval, denoted as $R=\{ r_{min}, r_{max}, I \} = \{1, 50, 20\}$. 
We replace $f(d)$ with $g(d)$ and vary $D \in \{ 6, 18, 31, 43 \}$, and the results are shown in Tab.~\ref{tab:multi_ablation}. 
Using $g(d)$, the best result is reached when $D=31$, which is lower than using $f(d|\tau=2)$ (-0.6$\%$ in average $m$AP). The results confirm that dealing with proposals with different duration by using anchors of different sizes is effective to acquire reliable proposal quality.

\subsection{Integrating BREM with More Methods}

In order to demonstrate the effectiveness of the proposed method, BREM is integrated with more TAD methods. The results are shown in Tab.~\ref{tab:thumos} and Tab.~\ref{tab:anet}. According to these results, BREM achieves consistent improvement regardless of TAD methods and the improvement is more significant when combining with weakly methods, e.g. A2Net.

\begin{table}[]
    \centering
    \caption{Comparison of state-of-the-art methods with and without BREM on THUMOS14. * indicates ours implementation.}
    \begin{tabular}{l|rrrrrr}
        \toprule
        Method  & 0.3 & 0.4 & 0.5 & 0.6 & 0.7 & Avg. \\
        \midrule
        A2Net* & 56.5  & 51.1  & 43.0 & 31.1 & 16.6 & 39.7  \\
        A2Net+BREM & 62.0  & 56.9  & 47.0 & 34.2 & 21.1 & 44.2 \\
        Base & 68.5  & 63.7  & 56.6 & 45.8 & 31.0 & 53.1  \\
        Base+BREM & 70.7  & 66.1  & 60.0 & 50.1 & 36.4 & 56.7  \\
        AF & 75.5  & 72.5  & 65.6 & 56.6 & 42.7 & 62.6  \\
        AF+BREM & {76.5} & {73.2} & {66.9} & {57.7} & {43.7} & {63.6} \\
        \bottomrule
    \end{tabular}
    \label{tab:thumos}
\end{table}

\begin{table}[]
    \centering
    \caption{Comparison of state-of-the-art methods with and without BREM on ActivityNet. * indicates ours implementation.}
    \begin{tabular}{l|rrrr}
        \toprule
        Method  & 0.5 & 0.75 & 0.95 & Avg. \\
        \midrule
        A2Net* & 43.1 & 28.6 & 4.9 & 28.0 \\
        A2Net+BREM & 46.0 & 31.0 & 5.4 & 30.2 \\
        Baseline & 52.4 & 34.3 & 5.6  & 33.6\\
        Baseline+BREM & 52.2 & 35.4 & 5.1  & 34.3   \\
        AF* & 53.6 & 35.9 & 7.3  & 35.2 \\
        AF+BREM & 52.8 & 36.6 & 6.9 & 35.5\\
        AF (TSP) & 54.1 & 36.3 & 7.7  & 36.0 \\
        AF+BREM (TSP) & 53.7 & 37.9 & 6.9 & 36.2 \\
        TCANet* & 51.7 & 36.3 & 10.3 & 35.5\\
        TCANet+BREM & 52.2 & 36.3 & 10.3 & 35.7\\
        \bottomrule
    \end{tabular}
    \label{tab:anet}
\end{table}

\begin{table}[]
    \centering
    \caption{Comparison of inference speed between our method and other methods.}
    \begin{tabular}{l|cc}
        \toprule
        Method & Speed (ms) & Memory (MB) \\
        \midrule
        AFSD & 63.1 & 1495\\
        Baseline & 45.2 & 1215\\
        Baseline+BREM & 55.6 & 1533\\
        ActionFormer & 2109.9 & 1971\\
        ActionFormer+BREM & 2180.4 & 2251\\
        \bottomrule
    \end{tabular}
    \label{tab:speed}
\end{table}

\subsection{Comparison on Inference Speed}

We provide a comparison of inference speed of different methods with and without BREM on THUMOS14. All results are tested on a video with 25.6s, 30fps and on a server with an NVIDIA Tesla V100 GPU. As Tab.~\ref{tab:speed} shown, Baseline+BREM acquires 12$\%$ relative speed improvement compared to AFSD, which is previous state-of-the-art method. And the additional memory usage of BREM is negligible because almost 80$\%$ of the memory is consumed by the backbone.
As for ActionFormer, because of time-consuming feature extraction  (98$\%$ of total time), the inference speed is lower than Baseline and BREM increases only 3.3$\%$ inference time. Above results show that BREM can bring considerable improvement with negligible memory and runtime cost.



\end{document}